%% file: conference_101719.tex
\documentclass[conference]{IEEEtran}
\IEEEoverridecommandlockouts
\usepackage{cite}
\usepackage{amsmath,amssymb,amsfonts}
\usepackage{algorithmic}
\usepackage{graphicx}
\usepackage{textcomp}
\usepackage{xcolor}

\usepackage{placeins}
\usepackage[linesnumbered,ruled,vlined]{algorithm2e}
\usepackage{subcaption}
\usepackage{graphicx}
\usepackage{multirow}
\usepackage{url}       

\SetCommentSty{mycommfont}

\SetKwInput{KwInput}{Input}                
\SetKwInput{KwOutput}{Output}              

\def\BibTeX{{\rm B\kern-.05em{\sc i\kern-.025em b}\kern-.08em
    T\kern-.1667em\lower.7ex\hbox{E}\kern-.125emX}}

\begin{document}
\newcommand{\method}{\texttt{AEBNAS}}
\title{AEBNAS: Strengthening Exit Branches in Early-Exit Networks through Hardware-Aware Neural Architecture Search\\
}

\author{
    \IEEEauthorblockN{
        \IEEEauthorrefmark{1}Oscar Robben, \IEEEauthorrefmark{1}Saeed Khalilian, Nirvana Meratnia
    }
    \IEEEauthorblockA{
       \textit{Eindhoven University of Technology}, Eindhoven, The Netherlands \\
        Email: \{o.a.m.robben, s.khalilian.gourtani, n.meratnia\}@tue.nl
    }
}
\renewcommand{\thefootnote}{\fnsymbol{footnote}}
\maketitle

\begin{abstract}
Early-exit networks are effective solutions for reducing the overall energy consumption and latency of deep learning models by adjusting computation based on the complexity of input data. By incorporating intermediate exit branches into the architecture, they provide less computation for simpler samples, which is particularly beneficial for resource-constrained devices where energy consumption is crucial. However, designing early-exit networks is a challenging and time-consuming process due to the need to balance efficiency and performance. Recent works have utilized Neural Architecture Search (NAS) to design more efficient early-exit networks, aiming to reduce average latency while improving model accuracy by determining the best positions and number of exit branches in the architecture. Another important factor affecting the efficiency and accuracy of early-exit networks is the depth and types of layers in the exit branches. In this paper, we use hardware-aware NAS to strengthen exit branches, considering both accuracy and efficiency during optimization. Our performance evaluation on the CIFAR-10, CIFAR-100, and SVHN datasets demonstrates that our proposed framework, which considers varying depths and layers for exit branches along with adaptive threshold tuning, designs early-exit networks that achieve higher accuracy with the same or lower average number of MACs compared to the state-of-the-art approaches.
\end{abstract}

\begin{IEEEkeywords}
Early-Exit Networks, Hardware-Aware NAS, Efficient Deep Learning, Dynamic Neural Networks
\end{IEEEkeywords}

\section{Introduction}
\footnotetext[1]{First two authors contributed equally.}
Deep neural networks have proved to be viable and effective solutions in a wide range of Internet of Things (IoT) applications, such as smart healthcare \cite{ahmed2021deep}, smart homes \cite{manu2019smart}, and smart agriculture \cite{zhu2018deep}. Deploying deep learning models on edge devices in IoT applications, rather than transferring data to the cloud for decision-making, offers advantages such as privacy preservation, low latency, and cost-effectiveness. Nevertheless, deploying deep learning models on edge devices is challenging due to their limited computational capacity and energy constraints. 

Numerous strategies, such as pruning \cite{li2016pruning}, quantization \cite{balzer1991weight}, weight sharing \cite{han2015deep,khalilian2023escepe}, early-exiting \cite{teerapittayanon2016branchynet}, layer skipping \cite{wang2018skipnet}, and lightweight architectures \cite{wang2022lightweight}, have been proposed to enhance the accessibility of deep learning models on edge devices. Early-exiting is one of the effective approaches among these methods for significantly reducing the latency and computational cost of deep learning models \cite{zhang2021self}. Unlike static networks that offer uniform computation for all samples, early-exit networks enhance flexibility by incorporating exit branches that adjust computation levels according to the complexity of input data.

However, designing an early-exit network presents four major challenges: 1) determining the number of exit branches, 2) finding their positions within the model, 3) defining the exit policy, and 4) defining the architecture of the exit branches (including the number and types of layers). These factors can impact the trade-off between performance and efficiency of the early-exit models. Previous studies have manually designed early-exit networks \cite{teerapittayanon2016branchynet, zhang2021self}. This manual design process and optimization of the trade-off between performance and latency is time-consuming and challenging. Consequently, recent studies have employed Neural Architecture Search (NAS) to automate the process \cite{gambella2023edanas,gambella2024nachos}. For instance, EDANAS \cite{gambella2023edanas} explored automatic placement and the number of exit branches within the candidate backbone architecture, aiming to balance accuracy with the average number of Multiply-Accumulate (MAC) operations during the NAS optimization process. However, they employed a fixed architecture for the exit branches.

In this paper, we aim to address a significant challenge in designing early-exit networks, i.e., determining the number of layers for each exit branch and exploring different types and configurations of layers within these branches. To address these challenges, we utilize a multi-objective NAS framework \cite{lu2020nsganetv2} that takes into account both accuracy and latency during optimization. In addition to exploring the positions and numbers of exit branches as previously studied \cite{gambella2023edanas,gambella2024nachos}, we redesign the search space, incorporating dynamic exit threshold adjustment, and optimization goals of NAS. This modification enables the NAS search algorithm to explore various configurations of exit branches with different layers and architectures, aiming to design more efficient early-exit networks and to facilitate their deployment on resource-constrained devices. Our main contributions can be summarized as follows:
\begin{itemize}
\item We introduce a new NAS framework called~\method, which aims at designing more efficient early-exit architectures by searching both backbone architectures and the architectures of exit branches.
\item We incorporate a hyperparameter into the optimization objective that defines the target number of MACs for the search algorithm, enabling the design of early-exit networks with a maximum defined MAC and facilitating the design of optimal architectures based on the computational constraints of target devices.
\item After each NAS iteration, we perform an additional search to adjust the exit thresholds based on the target average number of MACs and accuracy, and incorporate the updated results into the search population.
\item We present a comparison with previous approaches on the CIFAR-10 \cite{cifar}, CIFAR-100 \cite{cifar}, and SVHN \cite{svhm} datasets and demonstrate that~\method~achieves lower average number of MACs while achieving comparable or higher levels of accuracy.
\end{itemize}

\begin{figure*}[!t]
    \includegraphics[width=\textwidth, height=8cm]{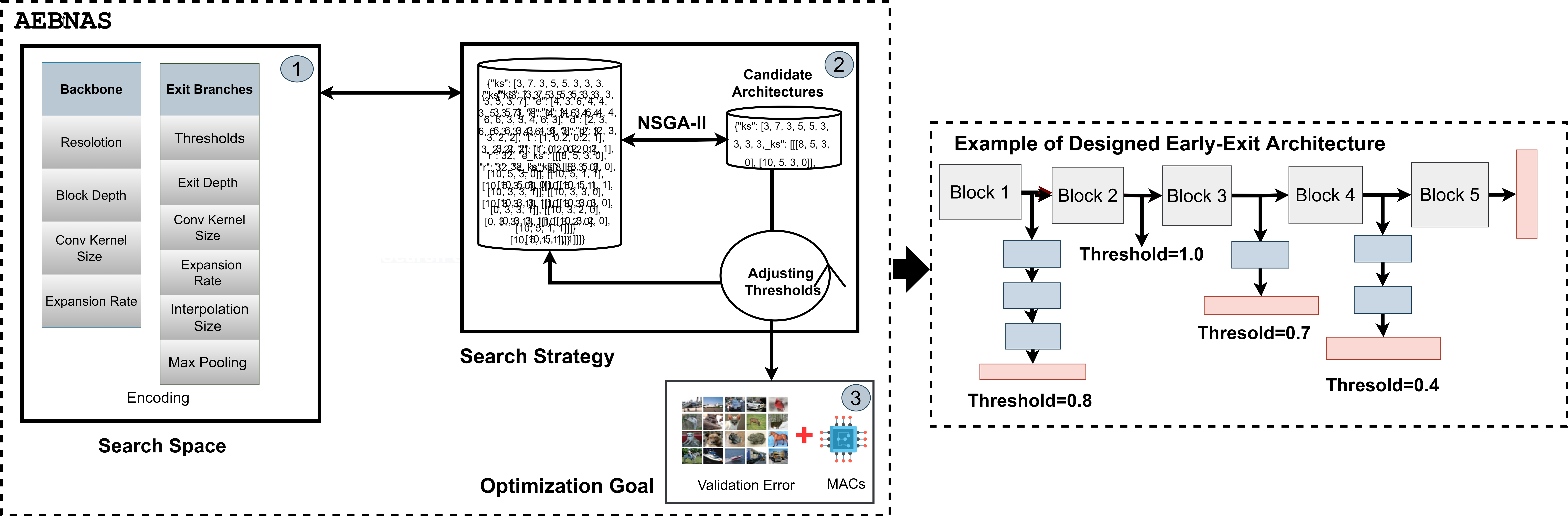}
    \label{fig:overview}
\caption{Illustration of the proposed NAS framework.~\method~includes three stages: 1) encoding early exit architectures in the search space, 2) tuning the exit thresholds, and 3) defining optimization goals for the search algorithm, balancing accuracy and the number of MACs for the designed early-exit architectures.}
\label{fig:overview}
\end{figure*}

\section{Related Works}
In this section, we review the most important approaches that have manually designed early-exit networks, as well as recent advancements that leverage NAS for designing and optimizing early-exit networks.

\subsection{Early-exit Networks}
Early-exit networks were first proposed by BranchyNet \cite{teerapittayanon2016branchynet}, where exit branches were manually added to the backbone architecture. The network, along with the exit layers, was trained through joint optimization. During inference, if the confidence of the predicted output at any exit layer was sufficient, the model halted further processing. This approach demonstrated that not all samples needed to be processed until the last layer, and by adding exit points, inference times were significantly reduced. Many approaches have since been proposed to improve performance and to reduce latency of early-exit methods by modifying the training or inference strategies.

Research by \cite{pmlr-v70-bolukbasi17a} introduced early-exiting using decision functions. After performing a certain number of convolutions, the decision function determined whether the sample should be evaluated further or exit the network. The optimization goal was to minimize inference time while keeping the error rate within a user-defined bound. The approach in \cite{guan2017energy} utilized the cascading of neural networks to assign samples that are easy to classify to classifiers that require less computation. This is a slight variation of early-exiting, where instead of having one main architecture with exit branches, the network is subdivided into separate classifiers. Each sample starts at the simplest classifier, and a stopping policy then decides whether the current prediction confidence is adequate or if the sample should be moved to the next classifier. The method described in \cite{8489276} utilized gate-classification layers to map activations to predictions. If the prediction confidence surpasses an automatically computed threshold, inference is terminated and the intermediate prediction is used as the final output of the network. The study in \cite{e24010001} utilized predictions of classification accuracies at different network exits. Their scheme required no hyperparameters and incorporated an exploration strategy for intermediate exits, leading to increased accuracy and reduced inference times.

ClassyNet \cite{10365527} employed early-exiting with a feature that ensures only specific classes of images exit at predefined layers of the network. EENet \cite{Ilhan_2024_WACV} introduces an early exit scheduler, where each exit is assigned a prediction and confidence score. EENet then calculates an exit score for each exit branch, terminating inference if the exit score exceeds a threshold. These thresholds are determined during the network's optimization phase, taking into account an inference budget.

Additionally, self-distillation \cite{zhang2021self} has demonstrated that strengthening the architecture of the exit branches leads to higher performance and faster inference. This approach showed that by leveraging knowledge distillation and attention modules, the performance of exit branches can be significantly enhanced \cite{zhang2021self}.

\subsection{Designing Early-exit Networks by NAS}

Due to the challenges in designing early-exit networks and finding the optimal trade-off between accuracy and efficiency, recent research has employed NAS to automatically design these networks.

HADAS \cite{bouzidi2023hadas} is a NAS framework that combines early exiting with dynamic voltage and frequency scaling to optimize both latency and energy consumption on edge devices. The framework searches for optimal positions and thresholds for exit branches. Similarly, EDANAS \cite{gambella2023edanas} focuses on identifying the optimal exit positions and thresholds but uses MACs as the hardware metric during the optimization process. However, EDANAS and HADAS keep a fixed architecture for the exit branches. NACHOS \cite{gambella2024nachos} imposes a constraint that each exit layer placed earlier in the network must have a lower computation time than layers placed later. This approach only allows the addition of max-pooling layers to the exit branches.

Inspired by previous works \cite{zhang2021self} that have demonstrated that strengthening exit branches can improve the performance and efficiency of models, we aim to address these challenges by enabling NAS frameworks to create exit branches with varying depths and architectures.

\section{Methodology}
We propose a new framework for designing early-exit networks by optimizing the variety of layers and the depth of exit branches to achieve a balance between efficiency and accuracy. Our proposed NAS framework, called~\method~is a multi-objective NAS framework built upon NSGANetV2 \cite{lu2020nsganetv2}, which was originally designed for creating static networks with a focus on improving accuracy and reducing latency. However, we design a new search space, adapt the optimization objectives, and introduce an additional phase for adjusting early-exit thresholds to enable the design of early-exit networks, as illustrated in Figure \ref{fig:overview}.

In this Section, we first describe the search space of~\method, detailing the encoding of early-exit networks. Next, we define our optimization objectives for the search algorithm, followed by the exit threshold tuning process using grid search. 


\subsection{Search Space}
The search space of~\method~comprises two main parts, i.e., (i) the encoded backbone architecture and (ii) the encoded exit branches. For the backbone architecture, we utilize the NSGANet \cite{lu2020nsganetv2} search space, which incorporates four important dimensions of convolutional neural networks and comprises five sequentially connected blocks. Within each block, the framework searches for the optimal number of layers, each employing an inverted bottleneck structure. The search process includes determining the expansion rate in the initial 1 × 1 convolution, the kernel size of the depth-wise separable convolution, the depth of each block, and the input resolution of the model.

To add the functionality of finding the best architecture for the exit branch, we encode the following key design elements in the search space: (i) the exit threshold, (ii) the number of exits, (iii) the positions of the exit branches, and (iv) the operations within the exit branches. The first three elements are represented by a single vector known as the \textit{threshold vector}. This vector stores the threshold value for each potential exit position, indicating that input samples with confidence levels exceeding the threshold can exit from that branch. A threshold value of 1 disables the exit at that position. This approach allows to control both the positions of the enabled exit branches and the total number of active exit branches. This encoding method is based on the EDANAS framework \cite{gambella2023edanas}.
 
The fourth element contains an array that specifies the operations for the exit layers. These layers follow a standard structure, starting with the interpolation of input activations, followed by a convolution layer, a batch-normalization layer, and a ReLU activation function. The associated hyperparameters of the convolution and batch normalization layers are included in the search space. These hyperparameters include kernel size, the expansion width of the convolutional layer, and interpolation size. Additionally, a max-pooling layer with a 2×2 kernel and a stride of 2 can optionally be placed after the ReLU function.

After the initial block, which includes the interpolation, convolutional layer, batch-normalization layer, ReLU function, and the optional max-pooling layer, a second block can be added that follows the same search space. Considering two possible values for each hyperparameter within an exit branch and two exit blocks, this configuration results in a total of 
$2^4 + 2^{4+4} = 272$  possible architectures for a single exit branch.

Sampling the initial population is done by randomly picking elements from the arrays storing the options for the interpolation size, kernel size and expansion rate. To determine whether a max-pooling layer is added or not we randomly pick either 0 or 1. Furthermore we have two arrays storing the interpolation size options, one for either block of the exit layer. The array storing the options for the second block also contains value 0 to allow the second block to be disabled. The sampled options are stored in two arrays for each block respectively. These two arrays are then stored in an array defining the all the exit-layers' configuration. The pseudo code of this process can be found in algorithm \ref{sampling}. 

\subsection{Optimization Goal}
Unlike static architectures with a fixed computational cost, early-exit networks have varying numbers of MACs depending on the input data. Therefore, the optimization process needs to consider the average computational cost. The average number of MACs is calculated by summing the MACs for all validation samples classified at different exit branches and dividing by the number of validation samples. In early-exit architectures, the average MAC count can vary significantly depending on the number of exit branches, their positions, and the depth and complexity of those branches. To control the maximum computational cost, we incorporate a hyperparameter into the optimization objective that sets the target number of MACs for the search algorithm. This enables the design of optimal architectures within the computational constraints of target devices. Our objective function is bi-objective, aiming to minimize both the average number of MACs and the model’s error rate, as defined in the equation below:

\begin{equation}
\label{eq:nas}
macs\_error(a) =  \frac{\left| predicted\_macs(a) - target\_macs \right|}{target\_macs} \nonumber
\end{equation}

\begin{gather}
\label{optgoal}
\small \min\  \Bigl( predicted\_error(a) + \beta \times macs\_error(a),\  \\ macs\_error(a)  \Bigr) \nonumber  \\  
\text{s.t. } a \in \mathcal{A} \nonumber
\end{gather}

, where $a$ is a network architecture within the set of candidate architectures $\mathcal{A}$, $predicted\_macs(a)$ and $predicted\_error(a)$ denote the predicted average number of MACs and the predicted error by surrogate models for architecture $a$, respectively. The parameter $target\_macs$ is the maximum target MACs, while \(\beta\) is used to further encourage improvements in accuracy as the solutions approach the target number of MACs.

\subsection{Confidence Thresholds Tuning}
A commonly used exit policy is applied during inference, where early exiting is controlled by a confidence-based metric known as the score margin. We compute this margin as the difference between the top-1 and top-2 softmax probabilities. The margin reflects model’s confidence in its prediction. Each exit branch is assigned a specific threshold, and the model exits early if the score margin exceeds this threshold, indicating sufficient confidence to terminate further computation. However, the selected thresholds have a significant impact on the average number of MACs. Setting lower threshold values for the first exit branches can substantially reduce the overall computational cost. To achieve an optimal trade-off between accuracy and efficiency, we perform a grid search after training each candidate architecture in every NAS iteration. This search identifies the optimal thresholds based on the following objective function. The updated thresholds, along with the evaluated architectures, are then used to train surrogate models that predict the accuracy and complexity of candidate models in the next NAS population.

\begin{equation}
\label{eq:grid_search}
\begin{split}
(1 - error)
&- \gamma \times 
   \frac{ \Biggl| 
      \frac{ \sum_{e=1}^{Exit} util(e)\cdot macs(e) }{N} 
      -target\_macs
   \Biggr| }
   {target\_macs}
\end{split}
\end{equation}

, where $error$ represents validation error of the architecture, $util(e)$ denotes number of samples exiting at branch $e$, $macs(e)$ is number of MACs required by branch $e$, and $N$ is total number of samples. The parameter $\gamma$ controls the trade-off between accuracy and efficiency.

\input{algorithm/sampling}

\section{Experiments}

\subsection{Setup}
\input{tables/setup_tb}
\input{tables/baselines}

\noindent \textbf{Datasets.}
We used the following datasets for performance evaluation: CIFAR-10 \cite{cifar}, CIFAR-100 \cite{cifar}, and SVHN \cite{svhm} datasets. For CIFAR-10 and CIFAR-100 datasets, the training set contains 50,000 images and the test set contains 10,000 images. In case of SVHN dataset, the training set consists of 73,257 images, while the test set includes 26,032 images. For all datasets, 10\% of the training set was used as a validation set.

\noindent \textbf{Hyperparameters.} All experiments were conducted using the hyperparameter values listed in Table \ref{table:setup}. To ensure a fair comparison with existing studies, target MAC values were set to match the range of models designed in previous works. Specifically, the targets were set to 1 million for SVHN dataset; 2 million and 17 million for CIFAR-10 dataset; and 17 million for CIFAR-100 dataset. For CIFAR-100 dataset, the baseline models, as well as NAS-based models in the last two iterations, were trained for 250 epochs. The exit branch hyperparameters were defined as follows: kernel sizes of 3 or 5, expansion widths of 1 or 2, interpolation sizes of 8, 10, or 12, and an optional max-pooling layer after each exit block. The maximum number of exit branch blocks was set to 2, while the maximum number of exit branches was limited to 5, with one allowed after each backbone block. The remaining NAS hyperparameters followed the settings of NSGANetV2.

\noindent \textbf{Profiling MACs.}
 We measured the MACs of the backbone networks using the TorchProfile MACs profiler\footnote{\url{https://github.com/zhijian-liu/torchprofile}}, which provides a reasonably accurate estimate of MACs of neural network operations. Following previous works that utilized TorchProfile, we adopted it for consistency in measuring the complexity of backbone architectures. However, since TorchProfile does not account for batch-normalization operations, it tends to underestimate the actual MACs. To measure the MACs of the exit layers, we instead employed the Thop profiler\footnote{\url{https://github.com/Lyken17/pytorch-OpCounter/tree/master/thop}}. Using these two separate profilers enabled us to isolate and evaluate the contribution of the newly added exit-layer architecture to the total MAC count independently of the backbone. In addition, for the manually designed models, the number of MACs was measured using the Thop profiler.

\noindent \textbf{Baselines.} 
We compared performance of \method with four state-of-the-art approaches, including manually designed architectures and NAS-designed architectures. Two manually designed models were considered: (i) EEAlexNet \cite{pomponi2021probabilistic} with 5 exit branches, and (ii) EEResNet20 \cite{pomponi2021probabilistic} with 10 exit branches.
As for the NAS-based baseline models, we employed the configurations reported in EDANAS \cite{gambella2023edanas} and NACHOS \cite{gambella2024nachos}. Both methods were configured with 30 search iterations, an initial population of 100 samples, and employed adaptive-switching surrogate models for predicting both MACs and validation error. EDANAS used 5 training epochs per architecture, while NACHOS increased the total training budget to 20 epochs, distributed over three stages of 10, 5, and 5 epochs. In our experiments, we followed the same NAS and training hyperparameter settings as EDANAS.

For each model, we report the average number of MACs, top-1 accuracy, the number of exit branches, and the utilization of each exit branch, which indicates the percentage of samples classified at each branch.

\subsection{Performance Comparison with Existing Baselines}
Table \ref{tb:baseline} presents the best-designed architecture from each baseline framework, as well as the best architecture discovered by~\method~, reporting their average number of MACs, the number of exit branches, top-1 accuracy, and the utilization of each exit branch.

For CIFAR-10 dataset, we report results targeting two average MACs budgets: 2 million and 17 million.~\method~designed early-exit models with 2.47 million MACs achieving 74.64\% accuracy. Compared to EDANAS and NACHOS at a similar MACs level,~\method~achieved significantly higher accuracy —6.86\% higher than EDANAS and 1.99\% higher than NACHOS. When compared to manually designed architectures,~\method~substantially reduced average number of MACs while achieving superior or comparable accuracy. At the higher target MACs,~\method~further improved accuracy, reaching 84.42\% with only 16.7 million MACs.

For SVHN dataset, with a target of 1 million MACs,~\method~achieved notably higher accuracy compared to EDANAS and NACHOS — improving by 6.04\% over EDANAS and 4.06\% over NACHOS — while maintaining a lower MACs count. Moreover, compared to manually designed models like EEAlexNet and EEResNet20, NAS-based frameworks significantly reduced computational costs.

For the CIFAR-100 dataset, which is a more complex task, it can be seen that manually designed architectures achieve relatively low accuracy despite a high number of MACs. In contrast, ~\method~achieves a 1.96\% improvement in terms of accuracy compared to EDANAS while using slightly fewer MACs. These results indicate that strengthening the architecture of exit branches and tuning the exit thresholds can improve the accuracy of the exit branches, allowing more samples to be classified earlier and thus achieving lower overall MACs.

Additionally, analyzing the number of exit branches shows that for a simpler dataset like SVHN,~\method~activated more exit branches compared to a more complex dataset like CIFAR-100. This indicates that adding numerous early exit branches in the backbone can increase computational overhead and lower overall accuracy, as seen with EEResNet20 using 10 exit branches. In contrast,~\method~effectively balances the number and architecture of exit branches with the model’s accuracy and computational efficiency.

\begin{figure*}[t]
\centering
\begin{subfigure}[t]{0.325\textwidth}
    \centering
    \includegraphics[width=\linewidth ,height=5.5cm]{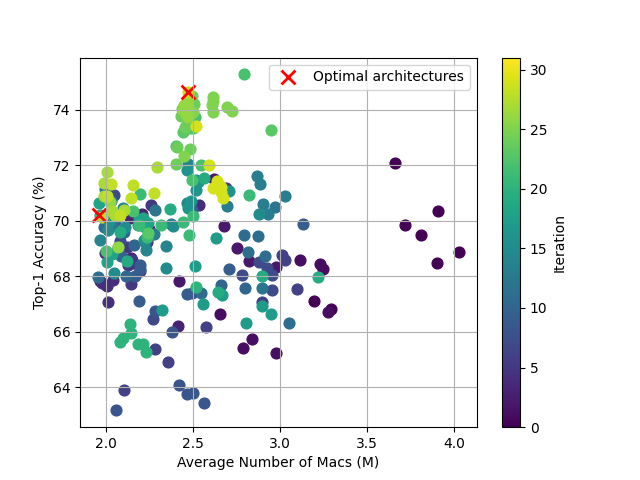}
    \caption{Evaluated architectures on CIFAR-10 dataset.}
    \label{fig:parento_plot_cifar} 
\end{subfigure}
\hfill
\begin{subfigure}[t]{0.325\textwidth}
    \centering
    \includegraphics[width=\linewidth,height=5.5cm]{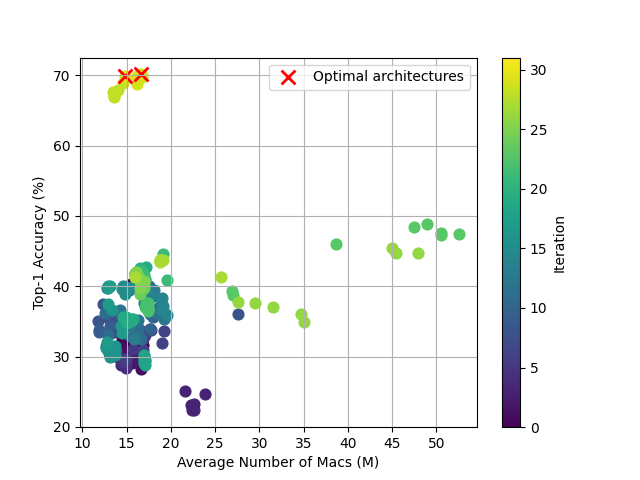}
    \caption{Evaluated architectures on CIFAR-100 dataset.}
    \label{fig:parento_plot_cifar100}
\end{subfigure}
\hfill
\begin{subfigure}[t]{0.325\textwidth}
    \centering
    \includegraphics[width=\linewidth,height=5.5cm]{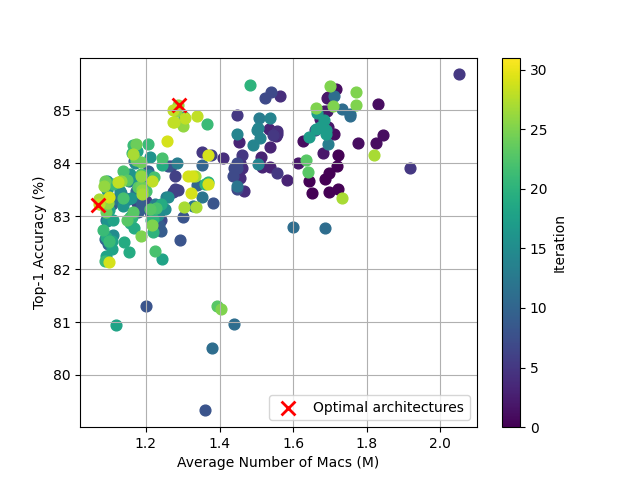}
    \caption{Evaluated architectures on SVHN dataset.}
    \label{fig:parento_plot_svhn}
\end{subfigure}
\caption{Comparison of top-1 accuracy (\%) and average number of MACs (in millions) for architectures assessed through multiple NAS iterations on (a) CIFAR-10, (b) CIFAR-100, and (c) SVHN datasets. Each point is color-coded by iteration number, with darker colors indicating earlier iterations and lighter colors representing later ones. Red crosses denote the optimal architectures identified during the search process.}
\label{fig:parento_plot}
\end{figure*}

\subsection{Accuracy vs. MACs Across NAS Iterations}
Following the predefined setup, NAS began by generating 100 random architectures, which were evaluated on the validation set to train surrogate models. In the subsequent 30 iterations, the system generated a total of 800 additional architectures but selected only 8 candidate designs per iteration based on the objective function for training and evaluation. In total, 340 architectures were assessed, 100 from the initial phase and 240 from the iterative process. Figure \ref{fig:parento_plot} illustrates the evolution of top-1 accuracy and average MACs across NAS iterations, with each data point color-coded by iteration: darker shades indicate earlier stages, while lighter tones correspond to later iterations.



Figure \ref{fig:parento_plot_svhn} presents all evaluated architectures on the SVHN dataset. The generated architectures cover a narrow range of top-1 accuracy and average MACs, indicating that the task is simpler compared to the CIFAR-10 and CIFAR-100 datasets. However, as the NAS process progressed, it was able to identify more compact architectures while maintaining a similar level of accuracy. 

Figure \ref{fig:parento_plot_cifar} shows the evaluated architectures on the CIFAR-10 dataset. Here, the generated architectures cover a wider range of both average MACs and top-1 accuracy. Notably, in the later iterations, the search process identified an effective trade-off between accuracy and computational cost, producing compact architectures that achieve higher accuracy. Figure \ref{fig:parento_plot_cifar100} presents the evaluated architectures on the CIFAR-100 dataset. The models in the last two NAS iterations were trained for a higher number of epochs, resulting in a notable jump in top-1 accuracy. Nevertheless, even when trained for only 5 epochs over 28 NAS iterations, NAS was able to improve top-1 accuracy by more than 40\%. After the 28th iteration, NAS identified architectures achieving around 50\% top-1 accuracy, while in the final iteration, architectures reached approximately 70\% top-1 accuracy with MACs close to the target of 17M.

Figure \ref{fig:parento_plot} demonstrates that integrating target MACs into the objective function and applying threshold tuning enables~\method~to converge successfully, producing architectures that meet target MACs with optimal accuracy.

\begin{figure}[h]
  \centering
  \includegraphics[width=\linewidth,height=8cm]{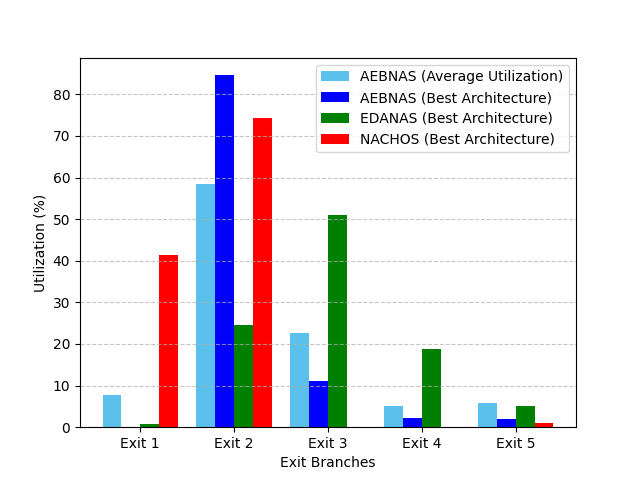}
  \caption{Analysis of the utilization of each exit branch (samples classified by each exit branch), including the average utilization during the search in~\method, along with the utilization of the optimal architectures identified by~\method, EDANAS, and NACHOS at a comparable MACs level (2.4M MACs).}
  \label{fig:exit_utilization}
\end{figure}

\subsection{Exit Utilization} 
Figure~\ref{fig:exit_utilization} shows the percentage of samples classified at each exit branch. It presents the average utilization pattern across all architectures evaluated by~\method~during the search, as well as the median exit utilization of the best architectures identified by~\method~, EDANAS, and NACHOS on the CIFAR-10 dataset at a computational cost of 2.4 million MACs.

Based on results presented in Table \ref{tb:baseline} and Figure~\ref{fig:exit_utilization}, analysis of the best architectures designed by~\method~shows that it disabled the first exit, classifying most samples through the second and third exit branches. By relying primarily on Exit 2, over 80\% of predictions could terminate early. In contrast, EDANAS and NACHOS used the first exit, which has lower performance and can negatively impact both overall accuracy and the model’s computational cost. Specifically, EDANAS activated the first exit, but only a few samples were classified there, creating additional computational overhead since all samples must pass through this exit to evaluate their confidence. This suggests that when an exit branch has low confidence, particularly in the early stages of the model, it may be beneficial to disable it.

Furthermore, analysis of the average utilization of ~\method~during the search shows that it predominantly generated architectures that actively used the second and third exit branches. These branches are located in the middle of the backbone architecture. Therefore, enhancing these branches can achieve a favorable balance between accuracy and computational cost. Additionally, comparing the utilization of the best architecture with the average utilization of~\method~during the search shows that the second exit branch was used 26\% more frequently in the best architecture than on average. In contrast, the other exit branches exhibited higher average utilization than in the best architecture. This indicates that the best architecture was discovered when the least effective exit branch (exit 1) was disabled, enabling most samples to be classified as early as possible at the stronger second exit branch.

\section{Conclusion}
In this paper, we presented a NAS-based framework,~\method, for generating early-exit neural networks. We encoded the backbone and exit branch architectures into the search space, allowing for the optimization of architectures specifically for the exit branches. By improving exit branch designs and tuning confidence thresholds during the search,~\method~generates more efficient early-exit models. Additionally, the system can be configured to target a specific average number of MACs, allowing the framework to generate models based on hardware requirements. A comparison with state-of-the-art techniques regarding accuracy and average number of MACs demonstrated that~\method~can generate models with higher accuracy while maintaining lower or comparable average MACs, highlighting the importance of including early-exit branch architectures in NAS search spaces.

\bibliographystyle{IEEEtran}  
\bibliography{references}     


\end{document}

%% file: algorithm/sampling.tex
\begin{algorithm}[!ht]
\DontPrintSemicolon
  
  \KwInput{$maxDepth, inter\_op, kern\_op, exp\_op$}
  \KwOutput{$exitlayers$}

  $mpool\_op := [0,1]$ \tcp*{Array specifying max-pooling option: enabled or disabled}
  $exitlayers[maxDepth] $ \tcp*{Initialize array storing the exit's configurations}

  $inter\_op\_b2 := [0]$ ++ $inter\_op$ \tcp*{Prepend 0 to the array storing the interpolation size options. This will be used to sample interpolation size of the second block of the exit-layer, 0 means this block is not enabled}

  \For{$i := 0$ \KwTo $maxDepth$}{
   \tcp*{Initialize arrays storing both configuration blocks}
    $block\_1[4]$\\
    $block\_2[4]$ \\

    \tcp*{Sample first configuration block}
    $block\_1[0] := $ randomChoice($inter\_op$) \\
    $block\_1[1] := $ randomChoice($kern\_op$) \\
    $block\_1[2] := $ randomChoice($exp\_op$) \\
    $block\_1[3] := $ randomChoice($inter\_op$) \\

     \tcp*{Sample second configuration block}
    $block\_2[0] := $ randomChoice($inter\_op\_b2$) \\
    $block\_2[1] := $ randomChoice($kern\_op$) \\
    $block\_2[2] := $ randomChoice($exp\_op$) \\
    $block\_2[3] := $ randomChoice($inter\_op$) \\

    $exitlayers[i] := [block\_1, block\_2]$   \tcp*{Store exit configuration in exit-layers array}
}

  \Return $exitlayers$

\caption{Early-Exit branch sampling algorithm}
\label{sampling}
\end{algorithm}

%% file: tables/setup_tb.tex
\begin{table}
\renewcommand{\arraystretch}{1.3} 
\scalebox{1.0}{
\centering
\begin{tabular}{lc}
\hline
\textbf{Parameter}                                       & \textbf{Value}        \\ \hline
Number of search iterations                              & 30                    \\
Number of training epochs                                & 5                     \\
Number of samples in the initial population              & 100                   \\
Number of samples in subsequent iterations               & 8                     \\
weight of the training loss                             & 1.0                      \\
supernet                                                &    MobileNetV3          \\
MACs and Error surrogate models                          & Multilayer perceptron \\
The \(\beta\) value in the equation \ref{eq:nas}  & 0.2                     \\
The \(\gamma\) value in the equation \ref{eq:grid_search} & 0.1        \\
Exit branch kernel sizes & 3, 5 \\
Exit branch expansion widths & 1, 2 \\
Exit branch interpolation sizes & 8, 10, 12 \\
Exit branch max-pooling options & 0, 1 \\
Maximum number of exit branch blocks & 2 \\
Maximum number of exit branches & 5 \\
Target MACs (M) value                                    & 1, 2, 17     \\ \hline
\end{tabular}
}
\caption{Parameter settings for our experiments.}
\label{table:setup}

\end{table}











%% file: tables/baselines.tex
\begin{table*}[h!]
\small
\begin{tabular}{lccccccc}
\hline
\textbf{Dataset}          & \textbf{Model}                 & \textbf{NAS} & \textbf{Exits}       & \textbf{MACs (M)}          & \textbf{Acc (\%)}                                        & \textbf{Utilization (\%)}                                                           \\ \hline
\multirow{9}{*}{CIFAR-10} &

              \textbf{~\method}        & yes          & 2      & \textbf{16.71}    & \textbf{84.42}      & [-, -, -, 93.38, 6.62] \\  
             & \textbf{~\method}        & yes          & 4      & \textbf{2.47}    & \textbf{74.64}     & [-,84.60, 11.20, 2.14, 2.06] \\
             & \textbf{~\method}        & yes          & 4      & 1.96   & 70.2      &  [-, 80.98, 11.02, 4.16, 3.84] \\

                          & EDANAS      & yes          & 2 & 17.31  & 80.90        & [-, -, -, 90, 10]\\
                           & EDANAS     & yes          & 5    & 2.47   & 67.78   & \scalebox{0.8}{[0.87 ± 1.17, 24.58 ± 7.56, 51.06 ± 12.65, 18.78 ± 10.17, 5.08 ± 4.65]}       \\
                          & NACHOS      & yes          & 4  & 2.44   & 72.65   & [41.31 ± 30.37, 74.31 ± 30.32, 0 , –, 0.90 ± 1.52] \\
                          & EEAlexNet   & no           & 5    & 233.46   & 73.03            & [38, 27, 19, 4, 12] \\
                          & EEResNet20  & no           & 10      & 21.25       & 69.71        & [6, 0, 0, 33, 16, 18, 8, 9, 6, 4]       \\ \hline
\multirow{7}{*}{SVHN} 
            & \textbf{~\method}       & yes          & 4   &   1.79   & 85.76         & [-,92.90, 4.74, 0.98, 1.38] \\
             & \textbf{~\method}       & yes          & 5  &    \textbf{1.10}    & \textbf{84.02}     &  [91.16, 6.68, 1.38, 0.24, 0.54]     \\
                          
                          & EDANAS     & yes           & 5  &   1.47  & 77.98         & \scalebox{0.8}{[24.97 ± 16.69, 82.94 ± 20.04, 1.28 ± 2.64, 0.24 ± 0.43, 0.56 ± 0.55]}   \\
                          & NACHOS      & yes          & 4  & 1.46   & 79.96     & [58.78 ± 21.89, 39.91 ± 24.51, 0, –, 1.31 ± 2.92]                  \\
                          & EEAlexnet  & no            & 5  & 263.51 & 89.00    & [23, 34, 27, 10, 6] \\
                          & EEResnet20 & no            & 10 & 70.20  & 88.53   & [0, 0, 0, 34, 13, 34, 11, 2, 2, 4] \\ \hline

\multirow{4}{*}{CIFAR-100} 
               & \textbf{~\method}      & yes    & 2   & 16.67    &    70.12    &   [-, -, -, 83.96, 16.04]  \\
               & \textbf{~\method}       & yes    & 2   & \textbf{14.80 }   &    \textbf{69.90}     &   [-, -, -, 82.42, 17.58]  \\

                          & EDANAS     & yes    & 3   &    14.94   & 67.94                 &  [34.12, -, -, 60.82, 5.06]   \\
                          & EEAlexnet  & no     & 5   &   435.76   &  60.97        & [4.66, 4.38, 3.06, 7.16, 80.74] \\
                          & EEResnet20   & no   & 10  &   25.47   & 53.09      &  [20.60, 9.34, 6.43, 5.71, 1.09, 1, 0.71, 0.13, 0.62, 54.37 ] \\ \hline
                          
\end{tabular}
\caption{Comparison of top-1 accuracy (Acc) and average number of MACs (in millions) for early-exit networks designed by~\method~on the CIFAR-10, CIFAR-100 and SVHN datasets, compared with other frameworks. The target MACs were set to 1M for SVHN, 2M and 17M for CIFAR-10, and 17M for CIFAR-100 . For each method, the number of exit branches, and the utilization of each exit branch are reported. A “–” in the utilization column indicates that the corresponding exit is disabled.}
\label{tb:baseline}
\end{table*}

%% file: references.bib
@String{Computer = "{IEEE} Computer" }

@String{Springer = "Springer-Verlag" }

@inproceedings{gambella2023edanas,
  title={EDANAS: Adaptive Neural Architecture Search for Early Exit Neural Networks},
  author={Gambella, Matteo and Roveri, Manuel},
  booktitle={2023 International Joint Conference on Neural Networks (IJCNN)},
  pages={1--8},
  year={2023},
  organization={IEEE}
}

@inproceedings{bouzidi2023hadas,
  title={HADAS: Hardware-Aware Dynamic Neural Architecture Search for Edge Performance Scaling},
  author={Bouzidi, Halima and Odema, Mohanad and Ouarnoughi, Hamza and Al Faruque, Mohammad Abdullah and Niar, Smail},
  booktitle={2023 Design, Automation \& Test in Europe Conference \& Exhibition (DATE)},
  pages={1--6},
  year={2023},
  organization={IEEE}
}

@inproceedings{wang2018skipnet,
  title={Skipnet: Learning dynamic routing in convolutional networks},
  author={Wang, Xin and Yu, Fisher and Dou, Zi-Yi and Darrell, Trevor and Gonzalez, Joseph E},
  booktitle={Proceedings of the European Conference on Computer Vision (ECCV)},
  pages={409--424},
  year={2018}
}

@inproceedings{teerapittayanon2016branchynet,
  title={Branchynet: Fast inference via early exiting from deep neural networks},
  author={Teerapittayanon, Surat and McDanel, Bradley and Kung, Hsiang-Tsung},
  booktitle={2016 23rd international conference on pattern recognition (ICPR)},
  pages={2464--2469},
  year={2016},
  organization={IEEE}
}

@ARTICLE{10365527,
  author={Ayyat, Mohammed and Nadeem, Tamer and Krawczyk, Bartosz},
  journal={IEEE Internet of Things Journal}, 
  title={ClassyNet: Class-Aware Early Exit Neural Networks for Edge Devices}, 
  year={2023},
  volume={},
  number={},
  pages={1-1},
  doi={10.1109/JIOT.2023.3344120}}

@inproceedings{lu2020nsganetv2,
  title={Nsganetv2: Evolutionary multi-objective surrogate-assisted neural architecture search},
  author={Lu, Zhichao and Deb, Kalyanmoy and Goodman, Erik and Banzhaf, Wolfgang and Boddeti, Vishnu Naresh},
  booktitle={Computer Vision--ECCV 2020: 16th European Conference, Glasgow, UK, August 23--28, 2020, Proceedings, Part I 16},
  pages={35--51},
  year={2020},
  organization={Springer}
}

@article{cifar,
  title={Learning multiple layers of features from tiny images},
  author={Krizhevsky, Alex and Hinton, Geoffrey and others},
  year={2009},
  publisher={Toronto, ON, Canada}
}

@article{gambella2024nachos,
  title={NACHOS: Neural Architecture Search for Hardware Constrained Early Exit Neural Networks},
  author={Gambella, Matteo and Pomponi, Jary and Scardapane, Simone and Roveri, Manuel},
  journal={arXiv preprint arXiv:2401.13330},
  year={2024}
}

@article{ahmed2021deep,
  title={A deep-learning-based smart healthcare system for patient’s discomfort detection at the edge of internet of things},
  author={Ahmed, Imran and Jeon, Gwanggil and Piccialli, Francesco},
  journal={IEEE Internet of Things Journal},
  volume={8},
  number={13},
  pages={10318--10326},
  year={2021},
  publisher={IEEE}
}

@article{manu2019smart,
  title={Smart home automation using IoT and deep learning},
  author={Manu, Rishabh Dev and Kumar, Sourav and Snehashish, Sanchit and Rekha, KS},
  journal={International Research Journal of Engineering and Technology},
  volume={6},
  number={4},
  pages={1--4},
  year={2019}
}

@article{zhu2018deep,
  title={Deep learning for smart agriculture: Concepts, tools, applications, and opportunities},
  author={Zhu, Nanyang and Liu, Xu and Liu, Ziqian and Hu, Kai and Wang, Yingkuan and Tan, Jinglu and Huang, Min and Zhu, Qibing and Ji, Xunsheng and Jiang, Yongnian and others},
  journal={International Journal of Agricultural and Biological Engineering},
  volume={11},
  number={4},
  pages={32--44},
  year={2018}
}

@article{li2016pruning,
  title={Pruning filters for efficient convnets},
  author={Li, Hao and Kadav, Asim and Durdanovic, Igor and Samet, Hanan and Graf, Hans Peter},
  journal={arXiv preprint arXiv:1608.08710},
  year={2016}
}

@article{balzer1991weight,
  title={Weight quantization in Boltzmann machines},
  author={Balzer, Wolfgang and Takahashi, Masanobu and Ohta, Jun and Kyuma, Kazuo},
  journal={Neural Networks},
  volume={4},
  number={3},
  pages={405--409},
  year={1991},
  publisher={Elsevier}
}

@article{zhang2021self,
  title={Self-distillation: Towards efficient and compact neural networks},
  author={Zhang, Linfeng and Bao, Chenglong and Ma, Kaisheng},
  journal={IEEE Transactions on Pattern Analysis and Machine Intelligence},
  volume={44},
  number={8},
  pages={4388--4403},
  year={2021},
  publisher={IEEE}
}

@article{han2015deep,
  title={Deep compression: Compressing deep neural networks with pruning, trained quantization and huffman coding},
  author={Han, Song and Mao, Huizi and Dally, William J},
  journal={arXiv preprint arXiv:1510.00149},
  year={2015}
}

@article{wang2022lightweight,
  title={Lightweight deep learning: An overview},
  author={Wang, Ching-Hao and Huang, Kang-Yang and Yao, Yi and Chen, Jun-Cheng and Shuai, Hong-Han and Cheng, Wen-Huang},
  journal={IEEE consumer electronics magazine},
  year={2022},
  publisher={IEEE}
}

@inproceedings{khalilian2023escepe,
  title={Escepe: Early-exit network section-wise model compression using self-distillation and weight clustering},
  author={Khalilian Gourtani, Saeed and Meratnia, Nirvana},
  booktitle={Proceedings of the 6th International Workshop on Edge Systems, Analytics and Networking},
  pages={48--53},
  year={2023}
}

@inproceedings{svhm,
  title={Reading digits in natural images with unsupervised feature learning},
  author={Netzer, Yuval and Wang, Tao and Coates, Adam and Bissacco, Alessandro and Wu, Baolin and Ng, Andrew Y and others},
  booktitle={NIPS workshop on deep learning and unsupervised feature learning},
  volume={2011},
  number={5},
  pages={7},
  year={2011},
  organization={Granada}
}

@article{pomponi2021probabilistic,
  title={A probabilistic re-intepretation of confidence scores in multi-exit models},
  author={Pomponi, Jary and Scardapane, Simone and Uncini, Aurelio},
  journal={Entropy},
  volume={24},
  number={1},
  pages={1},
  year={2021},
  publisher={MDPI}
}

@InProceedings{Ilhan_2024_WACV,
    author    = {Ilhan, Fatih and Chow, Ka-Ho and Hu, Sihao and Huang, Tiansheng and Tekin, Selim and Wei, Wenqi and Wu, Yanzhao and Lee, Myungjin and Kompella, Ramana and Latapie, Hugo and Liu, Gaowen and Liu, Ling},
    title     = {Adaptive Deep Neural Network Inference Optimization With EENet},
    booktitle = {Proceedings of the IEEE/CVF Winter Conference on Applications of Computer Vision (WACV)},
    month     = {January},
    year      = {2024},
    pages     = {1373-1382}
}

@article{guan2017energy,
  title={Energy-efficient amortized inference with cascaded deep classifiers},
  author={Guan, Jiaqi and Liu, Yang and Liu, Qiang and Peng, Jian},
  journal={arXiv preprint arXiv:1710.03368},
  year={2017}
}

@INPROCEEDINGS{8489276,
  author={Disabato, Simone and Roveri, Manuel},
  booktitle={2018 International Joint Conference on Neural Networks (IJCNN)}, 
  title={Reducing the Computation Load of Convolutional Neural Networks through Gate Classification}, 
  year={2018},
  volume={},
  number={},
  pages={1-8},
  doi={10.1109/IJCNN.2018.8489276}}

@InProceedings{pmlr-v70-bolukbasi17a,
  title = 	 {Adaptive Neural Networks for Efficient Inference},
  author =       {Tolga Bolukbasi and Joseph Wang and Ofer Dekel and Venkatesh Saligrama},
  booktitle = 	 {Proceedings of the 34th International Conference on Machine Learning},
  pages = 	 {527--536},
  year = 	 {2017},
  editor = 	 {Precup, Doina and Teh, Yee Whye},
  volume = 	 {70},
  series = 	 {Proceedings of Machine Learning Research},
  month = 	 {06--11 Aug},
  publisher =    {PMLR},
  url = 	 {https://proceedings.mlr.press/v70/bolukbasi17a.html}
}

@Article{e24010001,
AUTHOR = {Pomponi, Jary and Scardapane, Simone and Uncini, Aurelio},
TITLE = {A Probabilistic Re-Intepretation of Confidence Scores in Multi-Exit Models},
JOURNAL = {Entropy},
VOLUME = {24},
YEAR = {2022},
NUMBER = {1},
ARTICLE-NUMBER = {1},
URL = {https://www.mdpi.com/1099-4300/24/1/1},
PubMedID = {35052027},
ISSN = {1099-4300},
DOI = {10.3390/e24010001}
}
